\documentclass[12pt,journal,compsoc,onecolumn]{IEEEtran}
\usepackage{amsmath,graphicx,cleveref,caption,subcaption,dsfont,multirow,multicol,amssymb,amsfonts,mathtools,pifont}

 \usepackage{color}

\usepackage{booktabs}
\usepackage{lipsum}
\usepackage{bm}
\usepackage{stmaryrd}
\usepackage{algpseudocode}
\usepackage[ruled,vlined]{algorithm2e}
\usepackage[noadjust]{cite}
\let\OLDthebibliography\thebibliography
\renewcommand\thebibliography[1]{
  \OLDthebibliography{#1}
  \setlength{\parskip}{0pt}
  \setlength{\itemsep}{1.47pt plus 0.3ex}
}

\newcommand{\cmark}{\ding{51}}
\newcommand{\xmark}{\ding{55}}

\def\UU{{\bf U}}

\def\w{{\mathbf w}} 
 
\def\V{{\mathbf V}} 
\def\A{{\mathbf \Lambda}}

\def\A{{\bf A}}

\def\W{{\bf W}}
\def\WW{{\bf W}}
\def\SS{{\cal S}}

\def\N{{\cal N}}
\def\G{{\cal G}}
\def\V{{\cal V}}
\def\E{{\cal E}}
\def\F{{\cal F}}

\def\M{{\bf M}}

\title{Budget-Aware Graph Convolutional Network Design using Probabilistic Magnitude Pruning}
\author{ Hichem Sahbi \\ 
Sorbonne University, UPMC, CNRS, LIP6, France}

\begin{document}
\maketitle
\begin{abstract}
  Graph convolutional networks (GCNs) are nowadays becoming mainstream in solving many image processing tasks including skeleton-based recognition. Their general recipe consists in learning convolutional and attention layers that maximize classification performances.  With multi-head attention, GCNs are highly accurate but oversized, and their deployment on edge devices requires their pruning. Among existing methods, magnitude pruning (MP) is relatively effective but its design is clearly suboptimal as network topology selection and weight retraining are achieved independently. \\ In this paper, we devise a novel lightweight GCN design dubbed as Probabilistic Magnitude Pruning (PMP) that jointly trains network topology and weights. Our method is variational and proceeds by aligning the weight distribution of the learned networks with an a priori distribution. This allows implementing any fixed pruning rate, and also enhancing the generalization performances of the designed lightweight GCNs. Extensive experiments conducted on the challenging task of skeleton-based recognition show a substantial gain of our lightweight GCNs particularly at very high pruning regimes.  
 \end{abstract}
  \section{Introduction}
  \label{sec:intro}
  With the resurgence of deep neural networks~\cite{Krizhevsky2012}, many image processing and pattern recognition tasks have been successfully revisited during the last decade~\cite{He2016,He2017,HuangCVPR2017,Ronneberger2015}. These tasks have been approached with increasingly accurate but {\it oversized} networks, and this makes their deployment on cheap devices, endowed with limited hardware resources, highly challenging. Among existing models, graph convolutional networks (GCNs) are known to be effective particularly on non-euclidean domains including point-clouds and skeletons~\cite{Zhua2016,zhang2020}. Two categories of GCNs are known in the literature; spatial and spectral. Spectral methods \cite{kipf17,Levie2018,Li2018,Bresson16,Bruna2013,Henaff2015} first project graph signals from the input to the Fourier domain in order to achieve convolution \cite{spectral1997}, and then back-project the convolved signals in the input domain. Spatial methods \cite{Gori2005,Micheli2009,Wu2019,Hamilton2017} proceed differently by aggregating node signals using attention mechanisms prior to apply convolutions on the resulting node aggregates~\cite{attention2019}. Spatial GCNs  are deemed more effective compared to spectral ones, but their main downside resides in the high computational complexity especially when using multi-head attention. \\
\indent A major challenge in deep learning is how to make these networks lightweight and frugal while maintaining their high accuracy \cite{DBLP:conf/cvpr/HuangLMW18,DBLP:conf/cvpr/SandlerHZZC18,DBLP:journals/corr/HowardZCKWWAA17,DBLP:conf/icml/TanL19}. In this regard, many existing works tackle the issue of lightweight network design including tensor decomposition~\cite{howard2019}, quantization~\cite{DBLP:journals/corr/HanMD15}, distillation~\cite{DBLP:journals/corr/HintonVD15,DBLP:conf/iclr/ZagoruykoK17,DBLP:journals/corr/RomeroBKCGB14,DBLP:conf/aaai/MirzadehFLLMG20,DBLP:conf/cvpr/ZhangXHL18,DBLP:conf/cvpr/AhnHDLD19} and pruning~\cite{DBLP:conf/nips/CunDS89,DBLP:conf/nips/HassibiS92,DBLP:conf/nips/HanPTD15}. In particular, pruning methods are highly effective. Their  principle consists in removing connections whose impact on the classification performances is the least noticeable. Two major categories of pruning techniques exist in the literature; structured~\cite{DBLP:conf/iclr/0022KDSG17,DBLP:conf/iccv/LiuLSHYZ17} and unstructured~\cite{DBLP:conf/nips/HanPTD15,DBLP:journals/corr/HanMD15}. The former consists in zeroing-out weights of entire filters or channels whilst the latter seeks to remove weights individually and independently. Whereas structured  methods produce computationally more efficient networks, they are less effective compared to unstructured techniques; indeed, the latter provide more flexible (and thereby more accurate) networks which are  computationally still efficient. \\
\indent Magnitude pruning (MP) is one of the mainstream methods that proceeds by removing the smallest weight connections in a given heavy network, prior to retrain the resulting pruned (lightweight) network.  While being able to reach any targeted pruning rate, MP is clearly suboptimal as its design {\it decouples} the training of network topology from weights. Therefore, any removed connection cannot be recovered when retraining the pruned network, and this usually leads to a significant drop in classification performances. In this paper, we investigate a novel alternative for magnitude pruning referred to as PMP (Probabilistic Magnitude Pruning) that allows {\it coupling} the training of network topology and weights. The proposed method constrains the distribution of the learned weights to match any arbitrary targeted distribution and this allows, {\it via a band-stop  mechanism}, to filter out all the connections up to a given targeted pruning rate. Hence, the advantage of the proposed contribution is twofold; on the one hand, it allows reaching any pruning rate almost exactly, and on the other hand, it constrains the learned weights to fit a targeted distribution and this leads to better generalization as reported in experiments. 
    \section{Graph convnets at a glance}\label{gcnglance}
Let $\SS=\{\G_i=(\V_i, \E_i)\}_i$ denote a collection of graphs with $\V_i$, $\E_i$ being respectively the nodes and the edges of $\G_i$. Each graph $\G_i$ (denoted for short as $\G=(\V, \E)$) is endowed with a signal $\{\phi(u) \in \mathbb{R}^s: \ u \in \V\}$ and associated with an adjacency matrix $\A$. GCNs aim at learning a set of $C$ filters $\F$ that define convolution on $n$ nodes of $\G$ (with $n=|\V|$) as $(\G \star \F)_\V = f\big(\A \  \UU^\top  \   \W\big)$, here $^\top$ stands for transpose,  $\UU \in \mathbb{R}^{s\times n}$  is the  graph signal, $\W \in \mathbb{R}^{s \times C}$  is the matrix of convolutional parameters corresponding to the $C$ filters and  $f(.)$ is a nonlinear activation applied entry-wise. In  $(\G \star \F)_\V$, the input signal $\UU$ is projected using $\A$ and this provides for each node $u$, the  aggregate set of its neighbors. Entries of $\A$ could be handcrafted or learned so  $(\G \star \F)_\V$  makes it possible to implement a convolutional block with two layers; the first one aggregates signals in $\N(\V)$ (sets of node neighbors) by multiplying $\UU$ with $\A$ while the second layer achieves convolution by multiplying the resulting aggregates with the $C$ filters in $\W$. Learning  multiple adjacency (also referred to as attention) matrices (denoted as $\{\A^k\}_{k=1}^K$) allows us to capture different contexts and graph topologies when achieving aggregation and convolution.  With multiple matrices $\{\A^k\}_k$ (and associated convolutional filter parameters $\{\W^k\}_k$),  $(\G \star \F)_\V$ is updated as $f\big(\sum_{k=1}^K \A^k   \UU^\top     \W^k\big)$. Stacking aggregation and convolutional layers, with multiple  matrices $\{\A^k\}_k$, makes GCNs accurate but heavy. We propose, in what follows, a method that makes our networks lightweight and still effective.

  \section{Lightweight GCN Design}
In the remainder of this paper,   we subsume a given GCN as a multi-layered neural network $g_\theta$  whose weights are defined as $\theta =  \left\{\WW^1,\dots, \WW^L \right\}$, with $L$ being its depth,  $\WW^\ell \in \mathbb{R}^{d_{\ell-1} \times d_{\ell}}$ its $\ell^\textrm{th}$  layer weight tensor, and $d_\ell$ the dimension of $\ell$. The output of a given layer  $\ell$ is defined as
$ \mathbf{\phi}^{\ell} = f_\ell({\WW^\ell}^\top \  \mathbf{\phi}^{\ell-1})$, $\ell \in \{2,\dots,L\}$,  being $f_\ell$ an activation function; without a loss of generality, we omit the bias in the definition of  $\mathbf{\phi}^{\ell}$.\\
\noindent Pruning consists in zeroing-out a subset of weights in $\theta$ by multiplying $\WW^\ell$ with a binary mask $\M^\ell \in \{ 0,1 \}^{d_{\ell-1} \times d_{\ell}}$. The binary entries of  $\M^\ell$ are set depending on whether the underlying layer connections are kept or removed, so $\mathbf{\phi}^{\ell} = f_\ell( (\M^\ell \odot \WW^\ell )^\top \ \mathbf{\phi}^{\ell-1} )$, here $\odot$ stands for the element-wise matrix product. In this definition, entries of the tensor $\{\M^\ell\}_\ell$ are set depending on the prominence of the underlying connections in $g_\theta$. However, such pruning suffers from several drawbacks. On the one hand, optimizing the discrete set of variables $\{\M^\ell\}_\ell$ is known to be highly combinatorial and intractable especially on large networks. On the other hand,  the total number of parameters $\{\M^\ell\}_\ell$, $\{\WW^\ell\}_\ell$ is twice the number of connections in $g_\theta$ and this increases training complexity  and may also lead to overfitting.

\subsection{Band-stop Weight Parametrization}
\begin{figure}[h]
\centering
\resizebox{1.0\columnwidth}{!}{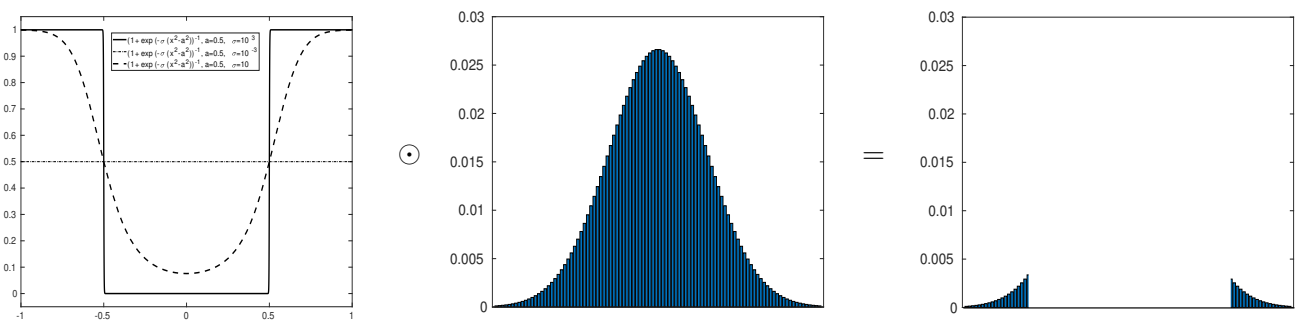}
\caption{This illustration shows a Band-stop function $\psi_{a,\sigma}$ and its application to a given (gaussian) weight distribution. Depending on the setting of $a$, only large magnitude weights are kept and correspond to the fixed pruning rate. (Better to zoom the file).}\label{tab2000}
\end{figure}

In order to circumvent the above issues, we consider an alternative {\it parametrization}, related to magnitude pruning, that allows finding both the topology of the pruned networks together with their weights, without doubling the size of the training parameters, while making learning still effective. This parametrization corresponds to the Hadamard product involving a weight tensor and a function applied entry-wise to the same tensor as
\begin{eqnarray}\label{eq2} \small
\WW^\ell = \hat{\WW}^\ell \odot \psi(\hat{\WW}^\ell),
\end{eqnarray}

\noindent here $\hat{\WW}^\ell$ is a latent tensor and $\psi(\hat{\WW}^\ell)$ is a continuous relaxation of $\M^\ell$ which enforces the prior that smallest weights should be removed from the network. In order to achieve this goal,  $\psi$ must be (i) bounded in $[0,1]$, (ii) differentiable, (iii) symmetric, and (iv) $\psi(\omega) \leadsto 1$ when $|\omega|$ is sufficiently large and $\psi(\omega) \leadsto 0$ otherwise. The first and the fourth properties ensure that the parametrization is neither acting as a scaling factor greater than one nor changing the sign of the latent weight, and also acts as the identity for sufficiently large weights, and as a contraction factor for small ones. The second property is necessary to ensure that $\psi$ has computable gradient  while the third condition guarantees that only the magnitudes of the latent weights matter. A possible choice, used in practice, that satisfies these four conditions is $\psi_{a,\sigma}(\hat{\w})=(1+\sigma \exp (a^2-\hat{\w}^2))^{-1}$ with $\sigma$ being a scaling factor and $a$ threshold. As shown in Fig.~\ref{tab2000}, $(\sigma, a)$ control the smoothness of $\psi_{a,\sigma}$ around the support $\Omega \subseteq \mathbb{R}$ of the latent weights. This allow implementing an annealed (soft) thresholding function that cuts-off all the connections in smooth and differentiable manner as training of the latent parameters evolves.  Put differently,  the asymptotic behavior of $\psi_{a,\sigma}$   --- that allows selecting the topology of the pruned subnetworks  --- is obtained as training reaches the latest epochs. 
\subsection{Probabilistic Magnitude Pruning}

The aforementioned parameterization --- while being effective (see later experiments) --- it does not allow to implement any targeted pruning rate as the dynamic of learned latent weights $\{\hat{\WW}^\ell\}_\ell$ is not known a priori. Hence, pruning rates could only be observed a posteriori or implemented after training using a two stage process (e.g., magnitude pruning + retraining). In order to implement any a priori targeted pruning rate as a part of a single  training process, we constrain the distribution of latent weights to fit an arbitrary probability distribution, so one may fix $a$ in $\psi_{a,\sigma}$ and thereby achieve the targeted pruning rate. Let $\hat{W} \in \Omega$ denote  a random variable standing for the latent weights in the pruned network $g_\theta$; $\hat{W}$ is assumed drawn from any arbitrary distribution $P$ (uniform, gaussian, laplace, etc). Fixing appropriately the distribution $P$ {\it not only} allows implementing any targeted pruning rate, but has also a regularization effect which controls the dynamic of the learned weights and thereby the generalization properties of  $g_\theta$ as shown subsequently and later in experiments.\\

\begin{figure}[h]
\centering
\resizebox{1.05\columnwidth}{!}{
 \hspace{-0.3cm} \includegraphics[width=0.33\linewidth]{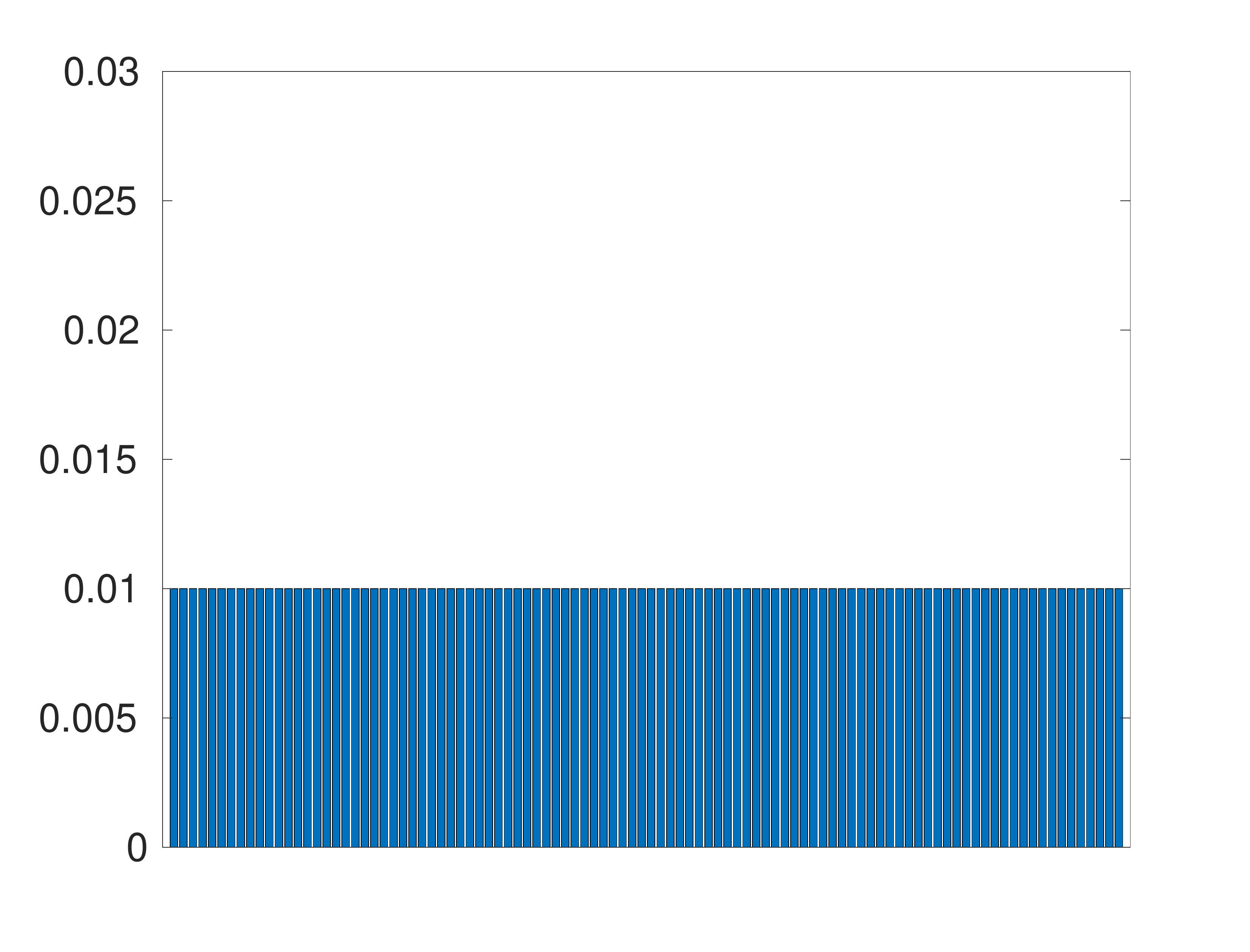} 
  \hspace{-0.3cm} \includegraphics[width=0.33\linewidth]{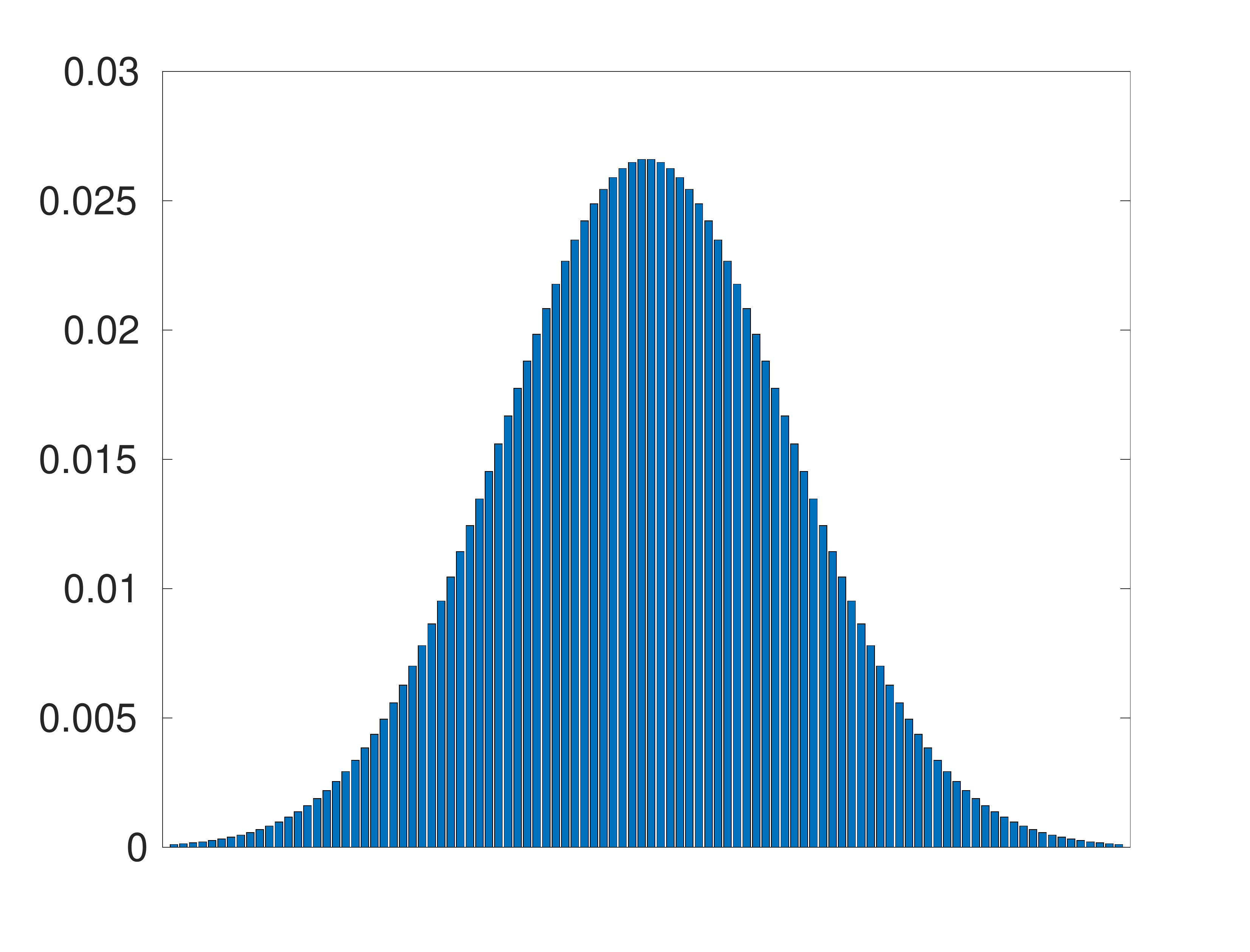}\hspace{-0.3cm} \includegraphics[width=0.33\linewidth]{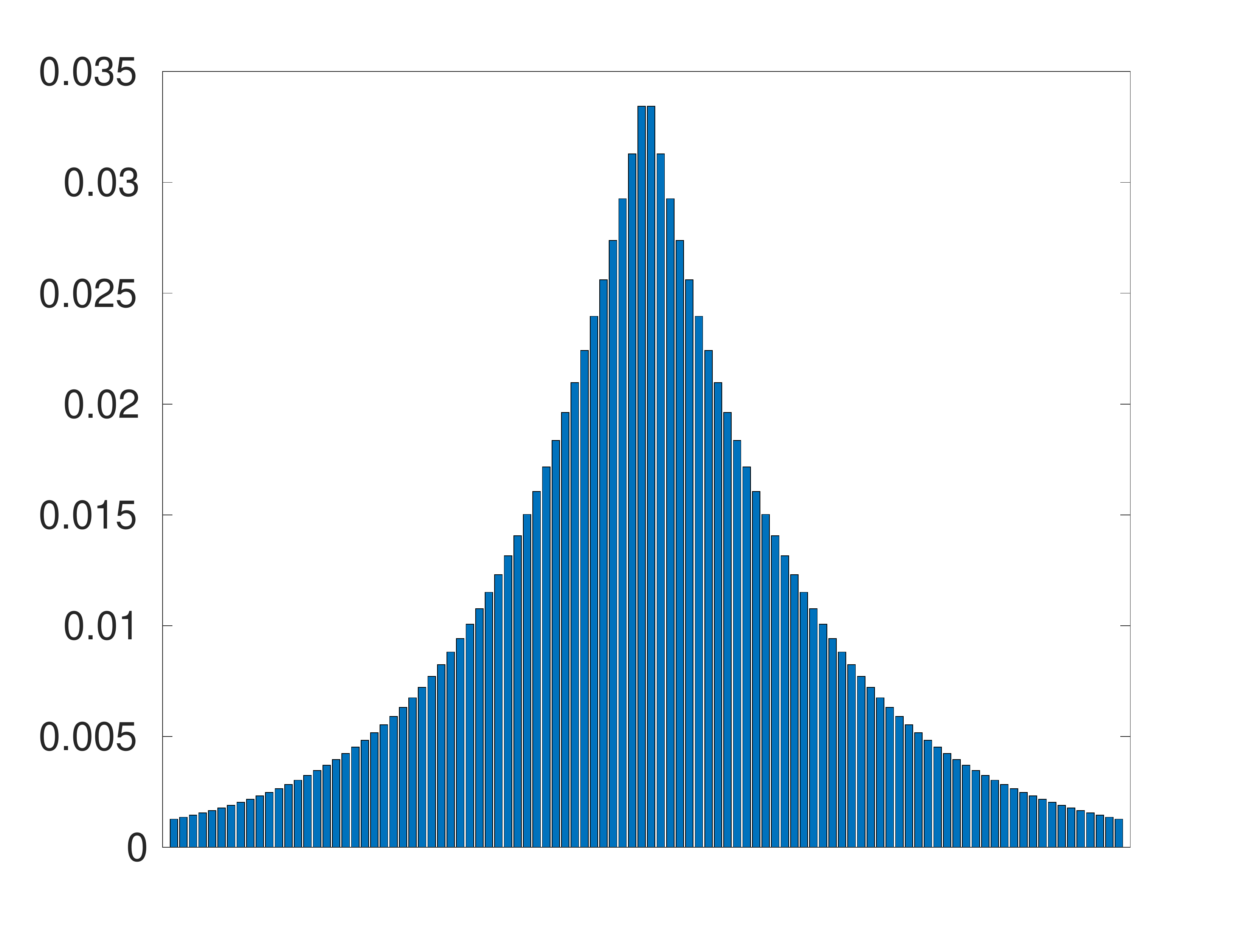} \hspace{-0.3cm}  \includegraphics[width=0.33\linewidth]{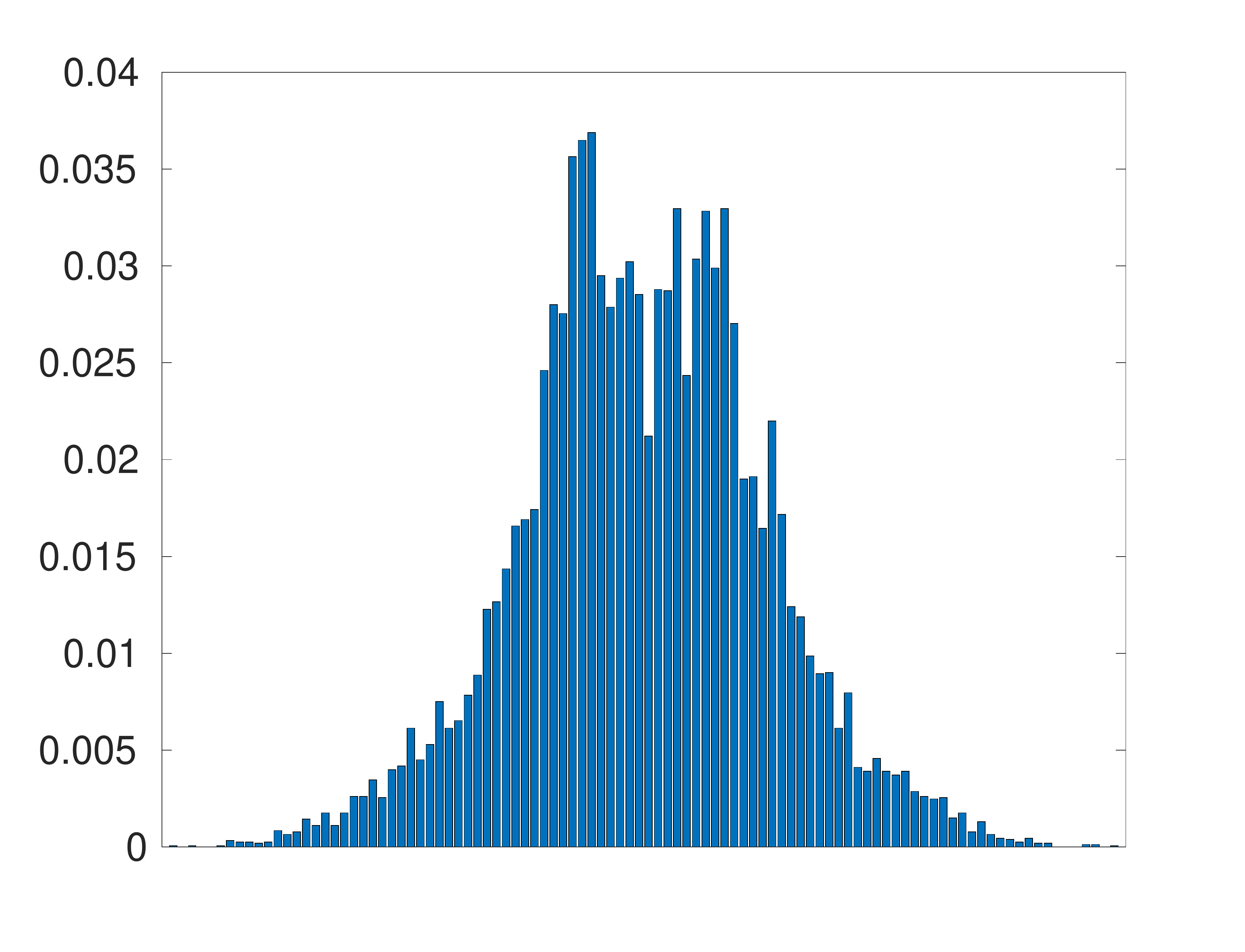}}
\caption{The first 3 figures correspond to targeted (uniform, gaussian and laplace) distributions. The 4th  figure shows the actual weight distribution of the heavy/unpruned GCN which resembles to gaussian. This may explain the best performances when the gaussian target is used particularly at low/mid pruning rates where the unpruned and pruned networks are more similar. At high pruning rates, laplace is better (see table~\ref{table21}).}\label{tab21}
\end{figure}

\noindent {\bf Fitting a targeted distribution.} Considering $Q$ as the observed distribution of the latent weights $\{\hat{\WW}^\ell\}_\ell$, and $P$ the targeted one, our goal is to reduce the discrepancy between $P$ and $Q$ using a Kullback-Leibler Divergence (KLD) loss
\begin{equation}\small
D_{KL}(P||Q)  = \int_{\Omega} P(\hat{W}) (\log P(\hat{W}) - \log Q(\hat{W})) \ d\hat{W}.
\end{equation}
Note that the analytic form of the above equation is known on the widely used probability density functions (PDFs), whilst for general (arbitrary) probability distributions, the exact form is not always known and requires sampling. Hence, we consider  instead a discrete variant of this loss as well as $P$ and $Q$; examples of discrete targeted distributions $P$ are given in Fig.~\ref{tab21} while the observed (and also differentiable) one $Q$ is based on a relaxed variant of histogram estimation. Let $\{q_1,\dots,q_K\}$ denote a $K$-bin quantization of $\Omega$ (in practice $K=100$), the k-th entry of $Q$ is defined as
\begin{equation}\label{eq3} \small
Q(\hat{W}=q_k) \propto \sum_{\ell=1}^{L-1} \sum_{i=1}^{n_\ell} \sum_{j=1}^{n_{\ell+1}} \exp\bigg\{-(\hat{\W}^\ell_{i,j}-q_k)^2/\beta_k^2\bigg\}, 
\end{equation}
\noindent here $\beta_k^2$ is a scaling factor that controls the smoothness of the exponential function; larger values of $\beta_k$ result into oversmoothed histogram estimation while a sufficiently (not very) small $\beta_k$ leads to a surrogate histogram estimation  close to the actual discrete distribution of $Q$. In practice, $\beta_k$ is set to $(q_{k+1}-q_k)/2$; with this setting, one may  replace $\propto$ (in Eq.~\ref{eq3})  with an equality as the partition function of $Q$ --- i.e.,  $\sum_{k=1}^K Q(\hat{W}=q_{k})$ ---  reaches almost one in practice.\\

\noindent {\bf Budget-aware pruning.}  Let $F_{\hat{W}}(a)=P(\hat{W}\leq a)$ be the cumulative distribution function (CDF) of $P(\hat{W})$. For any given pruning rate $r$, one may find the threshold $a$ of the parametrization $\psi_{a,\sigma}$ as
\begin{equation}\label{eq333} \small
a = F_{\hat{W}}^{-1}(r). 
 \end{equation}
 The above function, known as the quantile, defines the pruning threshold $a$ on the targeted distribution $P$ (and equivalently on the observed one $Q$ thanks to the KLD loss) which guarantees that only a fraction $(1-r)$ of the total weights are kept (i.e, nonzero) when applying the band-stop reparametrization in Eq.~\ref{eq2}. Note that the quantile at any given pruning rate $r$, can either be empirically evaluated on discrete random variables or can be analytically derived on the widely used PDFs (see table.~\ref{tab00}). \\

\noindent Considering the above budget implementation, pruning is achieved using a global loss as a combination of a cross-entropy term ${\cal L}_e$, and the KLD loss $D_{KL}$ (which controls weight distribution and hence guarantees the targeted pruning rate/budget depending on the setting of $a$ in $\psi_{a,\sigma}$ as shown in Eq.~\ref{eq333}) resulting into
\begin{equation}\label{eq34}\small
  \begin{array}{ll}
  \displaystyle  \min_{\{\hat{\WW}^\ell\}_\ell}  \displaystyle {\cal L}_e\big(\{\hat{\WW}^\ell \odot \psi(\hat{\WW}^\ell)\}_\ell\big)  \ + \ \lambda \ D_{KL}(P||Q), 
\end{array}
    \end{equation}
    \noindent    here $\lambda$ is sufficiently large (overestimated to $\lambda=10$ in practice), so Eq.~\ref{eq34} focuses  on implementing the budget and also constraining the pruning rate to reach $r$. As training evolves, $D_{KL}$  reaches its minimum and stabilizes while the gradient of the global loss becomes dominated by the gradient of ${\cal L}_e$, and this maximizes further the classification performances. \\
\begin{table}[hb]
\centering \resizebox{0.99\columnwidth}{!}{\begin{tabular}{c||c|c}
Distributions  & PDF $P(\hat{W})$ & Quantile $a=F_{\hat{W}}^{-1}(r)$ \\ 
  \hline
 \hline 
  Uniform   & $\frac{1}{T}$ & $a=\frac{r}{T}$ \\ 
  Gaussian  & $\frac{1}{\sigma \sqrt{2 \pi}} \exp\big\{-\frac{1}{2} \big(\frac{\hat{W}-\mu}{\sigma}\big)^2 \big\}$  & $a=\mu+\sigma \sqrt{2} \textrm{\bf erf}^{-1} (2r -1)$  \\ 
 Laplace &   $\frac{1}{2b} \exp\big\{-\frac{|\hat{W}-b|}{b} \big\}$ &  $a=\left\{\begin{array}{cc} \mu +b \ \log(2r) & \textrm{if  } r \leq \frac{1}{2} \\  & \\  \mu -b \ \log(2-2r) & \textrm{otherwise} \end{array}\right.$ 
\end{tabular}}
\caption{Different standard PDFs and the underlying quantile functions.}\label{tab00} 
\end{table}
\begin{table}
\resizebox{0.89\columnwidth}{!}{
\begin{tabular}{ccccc}
{\bf Method} & {\bf Color} & {\bf Depth} & {\bf Pose} & {$ \ \ \ \ \  \ \ \ \ \ \  \  $ \bf Accuracy (\%) $ \ \ \ \  \ \ \ \ \ \ \  \  $}\\
\hline
  Two stream-color \cite{refref10}   & \cmark  &  \xmark  & \xmark  &  61.56 \\
Two stream-flow \cite{refref10}     & \cmark  &  \xmark  & \xmark  &  69.91 \\ 
Two stream-all \cite{refref10}      & \cmark  & \xmark   & \xmark  &  75.30 \\
\hline 
HOG2-depth \cite{refref39}        & \xmark  & \cmark   & \xmark  &  59.83 \\    
HOG2-depth+pose \cite{refref39}   & \xmark  & \cmark   & \cmark  &  66.78 \\ 
HON4D \cite{refref40}               & \xmark  & \cmark   & \xmark  &  70.61 \\ 
Novel View \cite{refref41}          & \xmark  & \cmark   & \xmark  &  69.21  \\ 
\hline
1-layer LSTM \cite{Zhua2016}        & \xmark  & \xmark   & \cmark  &  78.73 \\
2-layer LSTM \cite{Zhua2016}        & \xmark  & \xmark   & \cmark  &  80.14 \\ 
\hline 
Moving Pose \cite{refref59}         & \xmark  & \xmark   & \cmark  &  56.34 \\ 
Lie Group \cite{Vemulapalli2014}    & \xmark  & \xmark   & \cmark  &  82.69 \\ 
HBRNN \cite{Du2015}                & \xmark  & \xmark   & \cmark  &  77.40 \\ 
Gram Matrix \cite{refref61}         & \xmark  & \xmark   & \cmark  &  85.39 \\ 
TF    \cite{refref11}               & \xmark  & \xmark   & \cmark  &  80.69 \\  
\hline 
JOULE-color \cite{refref18}         & \cmark  & \xmark   & \xmark  &  66.78 \\ 
JOULE-depth \cite{refref18}         & \xmark  & \cmark   & \xmark  &  60.17 \\ 
JOULE-pose \cite{refref18}         & \xmark  & \xmark   & \cmark  &  74.60 \\ 
JOULE-all \cite{refref18}           & \cmark  & \cmark   & \cmark  &  78.78 \\ 
\hline 
Huang et al. \cite{Huangcc2017}     & \xmark  & \xmark   & \cmark  &  84.35 \\ 
Huang et al. \cite{ref23}           & \xmark  & \xmark   & \cmark  &  77.57 \\  

  \hline
Our  GCN baseline                  & \xmark  & \xmark   & \cmark  &  \bf86.43                                                  
\end{tabular}}
\caption{Comparison of our baseline GCN against related work on FPHA.}\label{compare2}
 \end{table} 
\noindent Note that the impact of $D_{KL}(P||Q)$ in Eq.~\ref{eq34} has some similarities and differences w.r.t. the usual regularizers particularly $\ell_0$, $\ell_1$ and  $\ell_2$. Whilst these three regularizers favor respectively uniform, laplace and gaussian distributions in $Q$, there is no guarantee that $Q$ will {\it exactly match} an a priori distribution, so implementing any targeted pruning rate will require adding explicit (and difficult to solve) budget criteria. In contrast, as $Q$ is constrained in $D_{KL}(P||Q)$, the Band-pass  mechanism in Eq.~\ref{eq2} makes reaching any targeted pruning rate easily feasible. Note also that this Band-pass mechanism allows implementing a {\it partial} weight ranking --- through the $K$-bins of the distribution $Q$ --- in a differentiable manner. In other words, as training evolves, this approach jointly trains network topology $\psi_{a,\sigma}(\hat{\WW}^\ell)$ and weights $\hat{\WW}^\ell$  by (i) changing the {\it bin assignment} of $\hat{\WW}^\ell$ in $Q$, and by (ii) {\it activating and deactivating} these weights through $\psi_{a,\sigma}$ while maximizing generalization and satisfying exactly the targeted budget.

 \begin{table}
 \begin{center}
\resizebox{0.89\columnwidth}{!}{
  \begin{tabular}{ccccc}    
   \rotatebox{25}{Fixed PR}  &  \rotatebox{25}{Observed PR + gap}  & \rotatebox{25}{Target PDFs.} &      \rotatebox{25}{Accuracy (\%)}   & \rotatebox{25}{Observation}  \\
 \hline
  \hline
    none  &  0.00 (0.00) &\xmark              &    \bf86.43  & Baseline GCN (BGCN)\\
\multirow{3}{*}{\rotatebox{0}{0\%}}              & 0.00 (0.00)  &\cmark           &                 86.26    & BGCN+Uniform   \\     
                                                & 0.00 (0.00) &\cmark           &                 \bf 86.60    & BGCN+Gaussian   \\     
                                                & 0.00 (0.00) & \cmark          &                  86.26    & BGCN+Laplace  \\     
    \hline
  \multirow{4}{*}{\rotatebox{0}{55\%}} &  55.00 (0.00)   &\xmark        & 87.82  &  MP   \\
                                         & 55.10 (0.10)  &\cmark                        &   87.82   & MP+Uniform         \\
                                         & 55.31 (0.31) &   \cmark                      &  \bf88.52     & MP+Gaussian        \\
                                         & 57.83 (2.83)  &   \cmark                    &  87.65  & MP+Laplace       \\
   
    \hline
  \multirow{4}{*}{\rotatebox{0}{80\%}} &  80.00 (0.00)  & \xmark       & 86.78  & MP     \\
                                         &77.74 (2.26)  & \cmark                           &    85.91     & MP+Uniform      \\
                                         &80.71 (0.71)  &   \cmark                    &   \bf87.47     & MP+Gaussian     \\
                                         &80.11 (0.11)  &    \cmark                    &  86.95  & MP+Laplace        \\
  \hline  \multirow{4}{*}{\rotatebox{0}{98\%}} & 98.00 (0.00)  &   \xmark     &   60.34    & MP  \\
                                         & 97.98 (0.02) & \cmark                         &   70.26      & MP+Uniform    \\
                                         & 97.97 (0.03) &   \cmark                       & 70.60  &  MP+Gaussian      \\
                                         & 97.90 (0.10) &    \cmark                     &  \bf70.80   & MP+Laplace        \\
  \hline  \multirow{4}{*}{\rotatebox{0}{99\%}} & 99.01 (0.01)   &  \xmark      &  46.26  & MP    \\
                                         & 98.67 (0.33) & \cmark                                 &   62.43    &MP+Uniform       \\
                                         & 98.72 (0.28) &    \cmark                               & 56.17  &  MP+Gaussian      \\
                                         & 98.86 (0.14) &     \cmark                              & \bf64.17  & MP+Laplace       
                                       
  \end{tabular}}
\end{center}
\caption{Detailed performances and ablation study, for different fixed and observed pruning rates, and for different targeted probability distributions. Here ``PR'' stands for pruning rate, and ``gap'' as the difference between fixed and observed pruning rates.}\label{table21}
\end{table}
\section{Experiments}
  We benchmark our GCNs on the task of action recognition using  the First-Person Hand Action (FPHA) dataset~\cite{garcia2018} which includes 1175 skeletons belonging to 45 action  categories. Each sequence of skeletons (video) is initially described with a  graph  $\G = (\V,\E)$ with each node $v_j \in \V$ corresponding to the $j$-th hand-joint trajectory (denoted as $\{\hat{p}_j^t\}_t$)  and an edge $(v_j, v_i) \in  \E$ exists iff the $j$-th and the $i$-th trajectories are spatially connected.  Each trajectory in $\G$ is processed  using {\it temporal chunking} \cite{sahbiICPR2020}: first, the total duration of a  sequence is split into $M$ evenly-sized temporal chunks ($M=32$ in practice), then the trajectory  coordinates  $\{\hat{p}_j^t\}_t$  are assigned to the $M$ chunks (depending on their time stamps) prior to concatenate the averages of these chunks. This produces the raw description (signal) of $v_j$. \\

  \noindent {\bf Implementation details and baseline GCN.} We trained the GCNs end-to-end using the Adam optimizer \cite{Adam2014} for 2,700 epochs  with a batch size equal to $600$, a momentum of $0.9$ and a global learning rate (denoted as $\nu(t)$)  inversely proportional to the speed of change of our global loss used to train our networks. When this speed increases (resp. decreases),   $\nu(t)$  decreases as $\nu(t) \leftarrow \nu(t-1) \times 0.99$ (resp. increases as $\nu(t) \leftarrow \nu(t-1) \slash 0.99$).  We use in our experiments a GeForce GTX 1070 GPU (with 8 GB memory) and  we evaluate the performances using the protocol proposed in~\cite{garcia2018} with 600 action sequences for training and 575 for testing, and we report the average accuracy over all the classes of actions. The architecture of our baseline GCN (taken from \cite{sahbiICPR2020}; see also section~\ref{gcnglance}) includes stacked 8-head attentions applied to skeleton graphs whose nodes are encoded with 16-channels, followed by convolutions of 32 filters,  and a dense fully connected layer as well as a final classification layer. In total, this initial network is relatively heavy (for a GCN). Nevertheless, this GCN is accurate compared to the related work on the FPHA benchmark as shown in Table.~\ref{compare2}. Considering this GCN baseline, our goal is to make it lightweight while maintaining its high accuracy.\\

\noindent {\bf GCN performances.} Table~\ref{table21} shows the performances of our baseline and lightweight GCNs. From these results, we observe the positive impact for different PDFs and for increasing pruning rates $r$; for mid $r$ values (i.e., $55\%$), MP+gaussian  overtakes all the other setting while for very high pruning regimes (i.e., $\geq 98 \%$), MP+laplace is the most performant. Note that lightweight GCNs with mid $r$ overtake the baseline; indeed, mid $r$ values produce subnetworks with already enough (a large number of) connections and having some of them removed from the baseline GCNs produces a well known regularization effect~\cite{dropconnect2013}. We also note that the fixed and the observed pruning rates are very similar; the quantile functions of the gaussian and laplace PDFs allow implementing {\it fine-steps} of the targeted pruning rates particularly when $r$ is large. In contrast, the quantile functions of the gaussian and laplace PDFs are coarse around mid $r$ values (i.e., $55\%$).
\section{Conclusion}
We introduce in this paper a novel lightweight GCN design based on probabilistic magnitude pruning. The strength of the proposed method resides in its ability to constrain the probability distribution of the learned GCNs to match an a priori  distribution and this allows implementing any given targeted pruning rate while also enhancing the generalization performances of the resulting GCNs. Experiments conducted on the challenging task of skeleton-based recognition shows a significant gain of our method.  As a future work, we are currently investigating the extension of the current approach to other networks and databases.

\end{document}